\documentclass{article}

\usepackage{algorithm}
\usepackage{algpseudocode}
\usepackage{graphicx}   
\usepackage{float}      
\usepackage{placeins} 
\usepackage{listings}
\usepackage{xcolor} 
\usepackage[utf8]{inputenc}  
\usepackage{graphicx}
\usepackage[english]{babel}

\usepackage[letterpaper,top=2cm,bottom=2cm,left=3cm,right=3cm,marginparwidth=1.75cm]{geometry}

\usepackage{amsmath}
\usepackage{graphicx}
\usepackage[colorlinks=true, allcolors=blue]{hyperref}
\usepackage[utf8]{inputenc}
\usepackage{hyperref}
\usepackage{url}
\usepackage{centernot}
\usepackage{geometry}

\title{Think Inside the JSON: Reinforcement Strategy for Strict LLM Schema Adherence}
\author{
  Bhavik Agarwal, Ishan Joshi, Viktoria Rojkova \\
  MasterControl AI Research \\
  \texttt{\{bagarwal, ijoshi, vrojkova\}@mastercontrol.com}
}
\date{}

\begin{document}
\maketitle

\vspace{1em}

\begin{center}
\href{https://huggingface.co/MasterControlAIML/DeepSeek-R1-Qwen2.5-1.5b-SFT-R1-JSON-Unstructured-To-Structured}%
{\textbf{Hugging Face Model}}

\vspace{0.5em}

\href{https://huggingface.co/datasets/MasterControlAIML/R1-Reasoning-Unstructured-To-Structured}%
{\textbf{R1-Reasoning-Unstructured-To-Structured Dataset}}

\vspace{0.5em}

\href{https://huggingface.co/datasets/MasterControlAIML/JSON-Unstructured-Structured}%
{\textbf{JSON-Unstructured-Structured Dataset}}
\end{center}

\maketitle

\begin{abstract}
In this paper, we address the challenge of enforcing strict schema adherence in large language model (LLM) generation by leveraging LLM reasoning capabilities. Building on the DeepSeek R1 reinforcement learning framework, our approach trains structured reasoning skills of a 1.5B parameter model through a novel pipeline that combines synthetic reasoning data set construction with custom reward functions under Group Relative Policy Optimization (GRPO). Specifically, we first perform R1 reinforcement learning on a 20K sample unstructured to structured data set, mirroring the original DeepSeek R1 methods, to establish core reasoning abilities. Subsequently, we performed supervised fine-tuning on a separate 10K reasoning sample dataset, focusing on refining schema adherence for downstream tasks. Despite the relatively modest training scope, requiring approximately 20 hours on an 8×H100 GPU cluster for GRPO training and 3 hours on 1xA100 for SFT, our model demonstrates robust performance in enforcing schema consistency. We compare our ThinkJSON approach against the original DeepSeek R1 (671B), distilled versions of DeepSeek R1 (Qwen-1.5B and Qwen-7B) and Gemini 2.0 Flash (70B), showcasing its effectiveness in real-world applications. Our results underscore the practical utility of a resource-efficient framework for schema-constrained text generation. 
\end{abstract}

\section{Introduction}

In the highly regulated domain of bio-manufacturing quality, there is a growing need to convert legacy production records into structured digital formats for compliance and analysis. Biomanufacturing has historically been 'steeped in a paper culture', and even incremental moves toward electronic batch records are significant steps in industry digitalization \cite{Labant2025}. A key prerequisite of this digital migration is schema adherence: AI systems, such as large language models (LLMs) used to transcribe or summarize production logs, must output data that fit a predefined schema exactly. Any deviation (missing fields, incorrect format) could violate data integrity standards and render the generated records unusable for regulatory compliance \cite{Liu2024}. This introduces a critical challenge: While modern LLMs are extraordinarily powerful in free-form text generation, ensuring that they produce strictly structured, schema-valid outputs is not trivial. 

LLMs by default generate text probabilistically, with no built-in guarantee of conforming to a given format \cite{Geng2025}. This unpredictability poses risks when structured output is required for machine consumption or auditing. Empirical studies have found that even state-of-the-art models can \textit{fail} to consistently follow format instructions – success rates in producing correct JSON, for example, can vary widely from 0\% to 100\% depending on the task complexity and model used \cite{Shorten2024}. Such inconsistency is problematic in any setting, but in regulated bio-manufacturing, an output that does not exactly match the schema (e.g., a misformatted timestamp or an extra delimiter) might lead to compliance issues or require costly manual correction. Developers report that substantial effort is spent on prompt tuning and post-processing to coerce LLMs into the desired format \cite{Souza2024}. From a user perspective, unreliable formatting undermines trust – constraints help prevent nonsense or hallucinated fields, thereby ensuring the output remains credible and useful \cite{Souza2024}. In short, structured output generation is both a technical and a governance challenge: the model must be reliable in content as well as form.

\section{Relevant Work}
Researchers and practitioners are exploring several approaches to address these challenges and enforce schema adherence in LLM outputs. Key strategies include:
\subsection{Supervised Fine-Tuning}
An LLM can be fine-tuned on domain-specific data with the required output schema, so it learns to produce the correct structure. Fine-tuning on curated input–output pairs (e.g., historical records mapped to structured entries) can significantly improve format fidelity \cite{Li2024}. However, this approach is resource-intensive – training large models on specialized data is complex and costly, often requiring techniques like low-rank adaptation to be feasible \cite{Li2024}. Fine-tuning also risks making the model too domain-specific or rigid outside the training distribution.

\subsection{Reinforcement Learning with Human Feedback (RLHF)}
 
RLHF has proven effective in aligning LLMs with human instructions and preferences \cite{Wang2025}. By training a model with feedback signals that reward correct adherence to the desired format, one can encourage structured outputs. Notably, the instruction-following abilities of models like ChatGPT/GPT-4 are largely attributed to such alignment techniques \cite{Wang2025}, enabling them to obey fine-grained formatting requests (e.g. “output as JSON”). In regulated settings, RLHF could incorporate compliance-specific criteria into the reward model. The downside is that RLHF requires extensive high-quality feedback data and careful reward design; even then, smaller open-source models often still lag behind in format obedience despite alignment efforts \cite{Wang2025}.

\subsection{Constraint-Based Decoding} 
Rather than relying on the model to \textit{choose} the right format, constraint-based methods \textit{force} compliance by integrating schema rules into the generation process. Techniques like grammar- or regex-guided decoding intercept the model’s token output, only allowing continuations that keep the output valid according to a formal schema \cite{Geng2025}, \cite{Dong2024}. This guarantees 100\% schema adherence by construction. Recent frameworks implement fast, non-invasive constrained decoding that can guide LLMs to produce, for example, JSON that matches a given schema exactly \cite{Li2024}. Industry adoption of these ideas is rising; for instance, OpenAI’s API now accepts developer-provided JSON schemas and assures that the model’s response will conform to them \cite{Geng2025}. The trade-off here is potential complexity in setup and slight inference latency overhead, as well as the challenge of designing schemas that are neither over- nor under-constraining. Nonetheless, when correctness is paramount, constrained decoding is a powerful approach.

\subsection{Prompt Engineering} 
The most accessible technique is to craft the input prompt in a way that strongly cues the desired structure. This can involve giving the model explicit formatting instructions, examples of correctly formatted output, or even “layout hints” in the prompt. A well-designed prompt can often induce a model to produce a nearly perfect structured output \cite{Li2024}. Prompt engineering requires no model training and can be iteratively refined. However, it demands significant manual effort and expertise, and even then does not \textit{guarantee} consistency \cite{Li2024}. Models may still err on edge cases or as the prompt complexity grows, and maintaining long, complex prompts (especially across different models or updates) can be cumbersome. In practice, prompt-based solutions might be combined with lightweight validation or post-processing in high-stakes applications.

\subsection{Hybrid Constraint-Based Decoding and Prompt Engineering} 
By embedding knowledge of the schema at the prompt level and using a specialized procedure to keep the generation on track (via tagging, iterative re-checks, or extra control tokens), hybrid systems achieve schema adherence more reliably than a vanilla LLM approach \cite{Willard2023}. This structured, schema-first method is key to guaranteeing the outputs are valid, parseable, and aligned with downstream consumption requirements. Schema acts as a blueprint for how the final text must be organized while controllable generation mechanism conditions the model’s decoding process on these schema constraints. Instead of free-form text generation, the model is guided to fill in required slots, adhere to the correct format, and avoid extraneous or malformed outputs \cite{Willard2023}.

Each of these approaches comes with effectiveness trade-offs, especially under the stringent demands of regulated industries. Fine-tuning and RLHF can deeply instill format compliance into a model but at high development cost and with less transparency. Prompt engineering is more flexible and avoid retraining, but it relies on the base model’s capacity to follow instructions. Constraint-based decoding offers hard guarantees on structure, appealing for compliance, though it requires integrating external constraint logic with the model’s output stream. The choice often depends on the specific use case and constraints – for instance, biomanufacturers must consider not only technical accuracy but also validation, auditing, and data governance. Ensuring that LLM-generated records are \textit{both} accurate in content and precise in format is vital to meet quality and regulatory standards. Recent work underlines that reliable structured generation remains an open challenge, calling for continued research into methods that can robustly align LLM outputs with predefined schemas \cite{Shorten2024}. 

\section{Method}

Although the strategies outlined above—ranging from prompt engineering to constraint-based decoding—can improve structured output, they often require specialized tooling or large-scale fine-tuning. In regulated domains such as bio-manufacturing, these approaches must also be cost-effective and robust. In this section, we describe a reasoning-driven methodology that leverages synthetic data construction and iterative LLM reasoning to ensure schema adherence with minimal overhead. Specifically, we demonstrate how to:
\begin{itemize}
\item \textbf{Build RL reasoning dataset}
\begin{description}
    \item Create synthetic unstructured and structured data \cite{AM2023},\cite{Wa2022} in tandem using controlled prompts and Qwen 14B/32B \cite{QW2024},
    \item Reverse-engineer how unstructured text can map onto an empty JSON schema by engaging a distilled DeepSeek R1 Qwen 32B \cite{DS2025} to explain—step by step—how each schema field is populated.
\end{description}

\begin{figure}[H]
  \centering
  \includegraphics[width=0.99\textwidth]{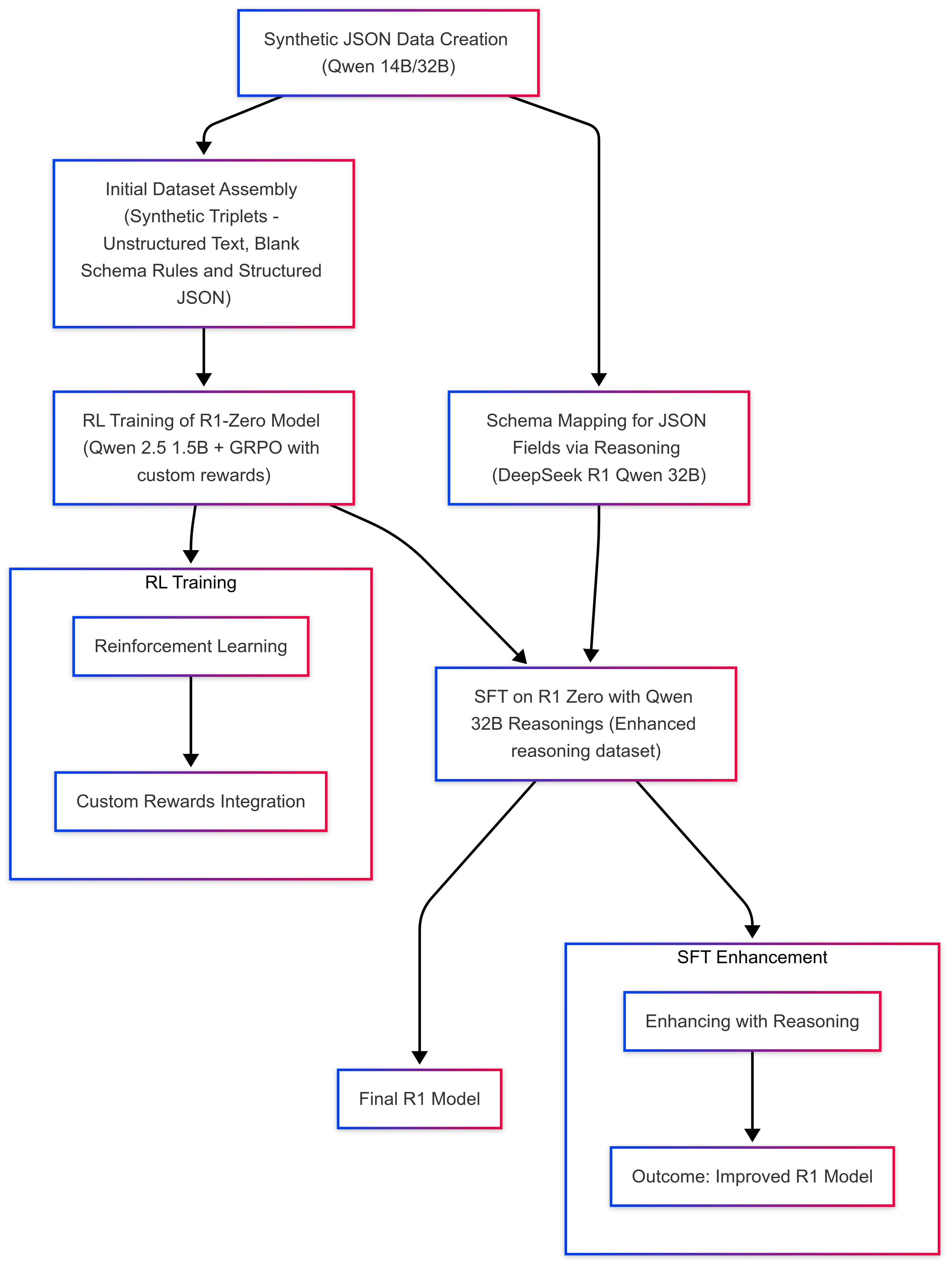}
  \caption{"Think inside the JSON" pipeline}
\end{figure}

\item \textbf {Train reasoning model with RL and SFT}. 
\begin{description}
    \item Develop custom reward mechanisms that directly evaluate how well the outputs adhere to a predefined schema while balancing fluency, completeness, and correctness.
    \item Train R1-Zero reasoning model from Qwen 2.5 1.5B base model using RL \cite{SH2024},\cite{W2023} and synthetic unstructured-structured pair dataset, integrate custom rewards into GRPO \cite{SH2024} without altering the core policy optimization loop. The combined reward drives the training so that the model produces outputs that score highly on all relevant criteria.
    \item Fine tune R1-Zero model into R1 with supervised fine-tuning using reasoning dataset.
\end{description}
\end{itemize}

\subsection{Generating Structured and Unstructured Data}

We begin by prompting a language model (Qwen 14B and 32B) to produce diverse, fully populated JSON schemas (including nested and complex fields). These filled schemas emulate real-world documentation (e.g., QA checklists, batch records) while showcasing variations in schema hardness and domain.
\begin{lstlisting}[language=TeX]
You are an expert in building a hierarchical JSON schema and object for the domain {DOMAIN}.
Your task is to create:

1. A multi-level JSON Schema describing:
   - ROOT (level 0),
   - SECTION (level 1),
   - SUBSECTION (level 2),
   - $DETAIL_{N}$ (level 3+).
   Each level may contain tables (2D data layouts) and checkbox elements (MCQs, confirmations),
   with nested components reflecting complex structures.

2. A JSON Object that strictly matches this schema, including:
   - \"id\" and \"title\"
   - \"level\" and \"$level_{type}$\"
   - An array of \"component\" objects (paragraphs, tables, or checkboxes)
   - A recursive \"children\" array
   - Special \"properties\" (e.g., \"variables\", \"content\") for data, logs, metrics, or formulas

Formatting Requirements:
- Escape all quotes (\"), replace newlines with \\n
- No trailing commas, single quotes, or extra data
- Enclose the final output with no extra explanations:
\end{lstlisting}

In parallel, we generate corresponding blank schemas—retaining structural outlines but omitting values. This gives us a “before and after” pair for each schema: an empty template and a filled instance. Such pairs are crucial for teaching LLMs how unstructured text should be systematically transformed into the exact JSON schema.
We then produce unstructured text reflecting the same content as the filled schema—but presented in varying layouts (e.g., sequential paragraphs, parallel sections, combined strategies) and table formats (ASCII art, XML/HTML-like snippets, simulated PDF extraction, etc.). These multi-format “narratives” mimic the real challenge of reading and interpreting inconsistent legacy documents.
\begin{lstlisting}[language=TeX]
You are an expert in generating hierarchical text documents from JSON Object data points.

**Task**: Convert the JSON Object into an unstructured, paragraph-based document.

**Given Data**: **Domain**; **JSON Schema**; **JSON Object**

**OUTPUT FORMAT** (enclosed strictly within <text>):
<text>
[Insert formatted hierarchical text from JSON object here]
</text>

**Layout References**:
- Layout options for components/levels: \n{RANDOM_LAYOUT}
- Table styles: \n{RANDOM_TABLE_STYLE}
- Checkbox styles: ```[ ], YES, NO, N/A, etc.```

**RULES**:
1. Map every JSON level, component, and attribute to the correct layout/style.
2. Surround JSON data points with additional words/sentences to obscure parsing.
3. Include all data (title, variables, metadata, content); no extra sections.
4. End each data point with a brief, unrelated remark.
5. Add filler paragraphs (definitions, domain info, etc.) not directly tied to the JSON content.
\end{lstlisting}
In doing so, we create a synthetic corpus that covers a broad range of domain contexts, from general manufacturing logs to specialized quality assurance frameworks. Each piece of unstructured text is logically equivalent to a filled JSON schema, yet differs in structure, style, and formatting. 

\subsection{Reasoning Dataset: Reverse-Engineering from Text to Schema}
We employ Distilled DeepSeekR1 Qwen 32B with the following prompt:
\begin{lstlisting}[language=TeX]
You are an AI assistant tasked with extracting structured data from a piece of text.

Inputs:
1. Text (source of information)
2. Blank Schema (unfilled JSON schema)
3. Filled Schema (final populated JSON)

Goals:
1. Compare Text + Blank Schema to the Filled Schema.
2. Explain step by step (chain-of-thought) how the text populates the blank schema.
3. Output only the reasoning steps (thought process).
4. Cross-verify that this reasoning exactly produces the Filled Schema.

Format your final response as:
Chain of Thought Explanation: """
\end{lstlisting}
The LLM is instructed to output only its chain-of-thought reasoning, explicitly describing the mapping from text to schema. Such self-explaining prompts push the model to maintain strict schema fidelity while revealing the logic behind each structural decision.
Because the prompt demands an explicit reasoning path, the LLM self-checks how each field is filled, minimizing random or malformed output. The chain-of-thought not only ensures correctness but also documents how the text was interpreted which is vital for regulated environments. By varying the domain (e.g., different types of QA reports) and text layout styles, we create a dataset that fosters LLM resilience to formatting quirks.

\subsection{GRPO Training on a Small Foundation Model}
Once we finalize the reasoning dataset, we proceed to train a small foundation model—mirroring the minimalistic DeepSeek R1 Zero approach—using GRPO \cite{DS2025}. We employ a 1.5B-parameter base model "to develop reasoning capabilities without any supervised data, focusing on their self-evolution through a pure reinforcement learning process" \cite{DS2025}. By leveraging a group-based advantage calculation and carefully designed reward signals (e.g., schema compliance, correctness), we efficiently instill structured reasoning capabilities within a resource-constrained pipeline.
By incorporating multiple reward functions \cite{CD2023} into the GRPO framework, we can simultaneously encourage format correctness (via r\_format) and content/domain correctness (via r\_equation). The combined reward drives training so that the model produces outputs that score highly on all relevant criteria. 
The entire process remains computationally light (e.g., ~20 hours on an 8×H100 cluster), demonstrating that strict schema adherence can be achieved even with compact, low-overhead foundation models.

\subsubsection{JSON-Based Reward}
This reward algorithm balances two aspects: (1) schema faithfulness via the key-value matching fraction, and (2) structural completeness via JSON length similarity. A high final reward indicates that the predicted JSON object closely matches the ground truth both in field contents and overall size.
\subsubsection{Format Verification Reward}
The format check enforces correct usage of specialized tags, crucial for downstream tasks that rely on clearly separated reasoning (\textless think\textgreater\ block) and final answers (\textless answer\textgreater\ block). The binary reward (0 or 1) simplifies reinforcement signals, focusing exclusively on structural correctness rather than content fidelity. The optional logging step enables sampling a small fraction of completions for qualitative inspection, aiding diagnostic or future training data curation.

\begin{algorithm}[H]
\caption{JSON-Based Reward Computation}
\label{alg:json-reward}
\begin{algorithmic}[1]

\State \textbf{Given:}
\Statex \quad A list of completions $\mathcal{C} = \{c_1,\dots,c_n\}$ from the model.
\Statex \quad A list of ground-truth JSON objects $\mathcal{G} = \{g_1,\dots,g_n\}$.
\Statex \quad Each $g_i$ is a valid JSON string.

\Procedure{ComputeReward}{$\mathcal{C}, \mathcal{G}$}
   \State $\mathcal{R} \gets \emptyset$   \Comment{Initialize rewards list}
   \For{\textbf{each pair} $(c_i, g_i)$ in $(\mathcal{C}, \mathcal{G})$}
       \State $c_i' \gets \texttt{"<think>"} \,\Vert\, c_i$ 
              \Comment{Insert \texttt{<think>} prefix}

       \State $ans \gets \text{substring}(c_i', \texttt{``<answer>''}, \texttt{``</answer>''})$
       \If{$ans$ is empty}
           \State $r_i \gets 0$ 
           \State \textbf{append} $r_i$ \textbf{to} $\mathcal{R}$; \textbf{continue}
       \EndIf

       \Comment{Parse as JSON}
       \State parse $ans$ into \texttt{answer\_json};
              parse $g_i$ into \texttt{gt\_json}
       \If{either parse fails}
           \State $r_i \gets 0$
           \State \textbf{append} $r_i$ \textbf{to} $\mathcal{R}$; \textbf{continue}
       \EndIf

       \Comment{Compute field overlap}
       \State $\mathcal{K}_a \gets \text{keys}(\texttt{answer\_json})$
       \State $\mathcal{K}_g \gets \text{keys}(\texttt{gt\_json})$
       \State $\textit{total\_fields} \gets |\mathcal{K}_a \cup \mathcal{K}_g|$
       \State $\textit{matching\_fields} 
              \gets \sum_{k \in (\mathcal{K}_a \cap \mathcal{K}_g)} 
                  \mathbf{1}[\texttt{answer\_json}[k] = \texttt{gt\_json}[k]]$

       \If{$\textit{total\_fields} > 0$}
           \State $\textit{key\_match\_score} \gets 
              \frac{\textit{matching\_fields}}{\textit{total\_fields}}$
       \Else
           \State $\textit{key\_match\_score} \gets 0$
       \EndIf

       \Comment{Compare JSON lengths}
       \State $\ell_a \gets \text{length}(\texttt{answer\_json}) \text{ or } 1$
       \State $\ell_g \gets \text{length}(\texttt{gt\_json}) \text{ or } 1$
       \State $\textit{length\_ratio} \gets \frac{\min(\ell_a,\ell_g)}{\max(\ell_a,\ell_g)}$

       \Comment{Calculate final reward}
       \State $r_i \gets \frac{\textit{key\_match\_score} + \textit{length\_ratio}}{2}$
       \State clamp $r_i$ to $[0,1]$; round to $1$ decimal place

       \If{$r_i \ge 0.6$}
           \State \textbf{log} $c_i'$ with $60\%$ probability
       \EndIf
       \State \textbf{append} $r_i$ \textbf{to} $\mathcal{R}$
   \EndFor
   \State \textbf{return} $\mathcal{R}$ 
\EndProcedure

\end{algorithmic}
\end{algorithm}

\begin{algorithm}[H]
\caption{Format Verification Reward}
\label{alg:format-verification}
\begin{algorithmic}[1]

\State \textbf{Goal:} Assign a reward of $0$ or $1$ depending on whether a generated completion follows an expected structure using \texttt{<think>}...\texttt{</think>} and \texttt{<answer>}...\texttt{</answer>}.
\vspace{0.5em}

\State \textbf{Inputs:}
\Statex \quad A list of completions $\mathcal{C} = \{c_1,\ldots,c_n\}$ (model-generated).
\Statex \quad A list of ground-truth objects $\mathcal{G} = \{g_1,\ldots,g_n\}$ (not directly used here, but included for extensibility).
\Statex \quad A small probability $p$ (e.g., $0.1$) for selectively logging completions.

\State \textbf{Output:} A list of scalar rewards $\mathcal{R} = \{r_1,\ldots,r_n\}$, with $r_i \in \{0,1\}.$
\vspace{0.5em}

\State Initialize an empty rewards list: $\mathcal{R} \gets [\,]$.

\For{\textbf{each pair} $(c_i, g_i)$ \textbf{in} $(\mathcal{C}, \mathcal{G})$}
  \State \textbf{Synthesize prompt format}:
   \quad $c_i' \gets \texttt{"<think>"} \,\Vert\, c_i$ 
              \Comment{Prepend ``\texttt{<think>}''}

  \State \textbf{Probabilistic logging}:
   \quad Draw $x \sim U[0,1]$. 
  \If{$x < p$}
    \State Log $c_i'$ to file for future analysis.
  \EndIf

  \State \textbf{Format check via regex}:
  \Statex \quad Define $\mathcal{R} =$ 
  \verb|"^<think>([^<]*(?:<(?!/?think>)[^<]*)*)</think>\n<answer>([\\s\\S]*?)</answer>$"|
  \Statex \quad Match $\mathcal{R}$ against $c_i'$. 
  \If{match fails (no correct grouping)}
    \State $r_i \gets 0$
  \Else
    \State $r_i \gets 1$
  \EndIf

  \State Append $r_i$ to $\mathcal{R}$.
\EndFor

\State \textbf{return} $\mathcal{R}$

\end{algorithmic}
\end{algorithm}

\begin{algorithm}[H]
\caption{GRPO with Multiple Reward Functions}
\label{alg:grpo-multi}
\begin{algorithmic}

\State \textbf{Notation and Setup} 
\Statex \quad Define a combined reward:
\[
  R_{\text{comb}}(c) \;=\; f\bigl(r_1(c),\,r_2(c),\,\ldots,\,r_K(c)\bigr),
\]
where \(f\) can be a weighted sum, mean, or any aggregator, \(\pi_\theta\) be the current policy (a language model parameterized by \(\theta\)), \(\{r_k\}_{k=1}^K\) be \(K\) reward functions (e.g., \(r_{\text{format}}\), \(r_{\text{equation}}\))

\vspace{0.5em}
\State \textbf{Group-Based Relative Advantage}
\Statex \quad Let \(G = \{\,c_1,\dots,c_M\}\) be a group of \(M\) outputs sampled from \(\pi_\theta\).
\Statex \quad For each \(c_i\), compute a combined reward \(R_i = R_{\text{comb}}(c_i)\).
\Statex \quad Define the relative (rank-based) advantage:
\[
   A^{(\text{rel})}(c_i) 
   \;=\; \frac{1}{M - 1}\;\sum_{j \neq i} 
         \mathbf{1}\bigl(R_i > R_j\bigr),
\]
which is the fraction of samples in \(G\) that have lower reward than \(c_i\).

\vspace{0.5em}
\State \textbf{GRPO Update}
\Statex \quad Update \(\theta\) to favor completions with higher relative advantage. 
\Statex \quad The GRPO loss for group \(G\) is:
\[
  \mathcal{L}_{\text{GRPO}}(\theta) 
    \;=\; -\,\sum_{c_i \in G} 
          A^{(\text{rel})}(c_i)\;\log \pi_\theta(c_i)
       \;+\;\text{Reg}(\theta),
\]
where \(\text{Reg}(\theta)\) includes regularization terms (e.g., entropy bonus, KL-divergence).

\end{algorithmic}
\end{algorithm}

\begin{figure}[H]
  \centering
  \includegraphics[width=0.99\textwidth]{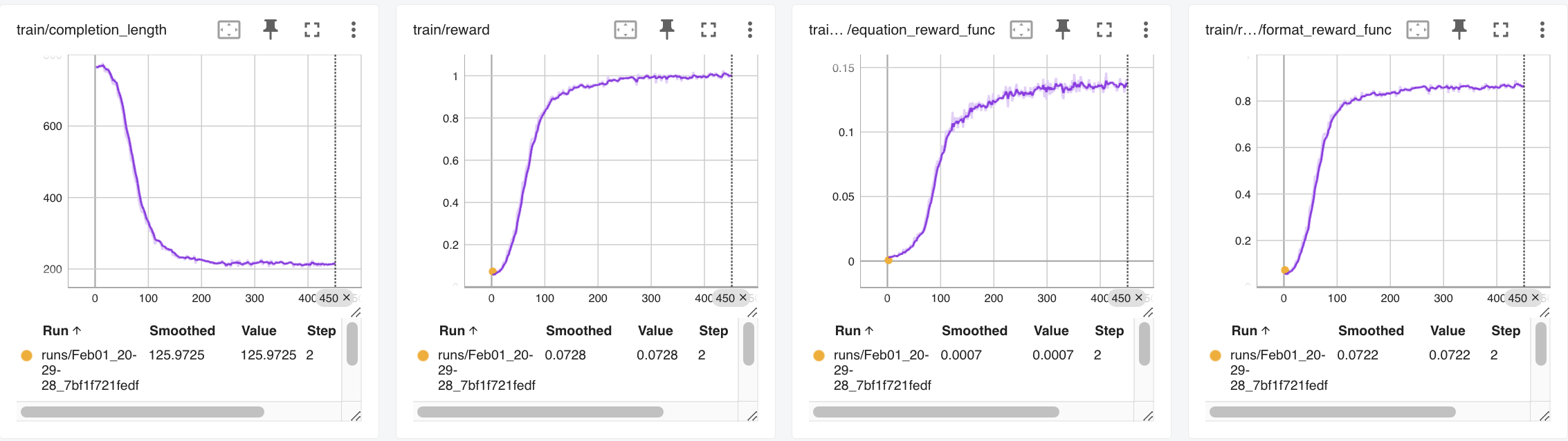}
  \caption{GRPO Training Metrics}
\end{figure}
\FloatBarrier
\subsection{Supervised Fine-Tuning}
While reinforcement learning confers advanced reasoning capacities, but supervised fine-tuning provides the final task- and schema-specific “polish” that ensures outputs are both logically grounded and robustly aligned with real-world standards. \cite{DS2025}. 
Reinforcement learning (RL) optimizes a policy for broad correctness or format adherence but can overlook rare or domain-specific intricacies (e.g., specialized field naming conventions, unusual data types). SFT exposes the model to explicit examples that emphasize precisely how to handle real-world edge cases, ensuring no field or condition is left under-represented.
Although RL fosters adaptability, the learned policy may still exhibit variability in ambiguous contexts or unrepresented task scenarios \cite{DS2025}. SFT, by contrast, anchors the final policy to concrete labeled examples, reducing output drift. 
By overlaying a final SFT stage, ThinkJSON tightly aligns its already-developed reasoning to the strict output requirements (e.g., correct JSON keys, mandatory fields), producing outputs suitable for audit or compliance.
For SFT (and SFT+LoRA) we used the Unsloth training framework on an A100 GPU, completing the process in about 3 hours.

\begin{lstlisting}[language=TeX]
### Role:
You are an expert data extractor mapping hierarchical text to a given JSON Schema.

### DATA INPUT: Text; Blank JSON Schema

### TASK REQUIREMENT:
1. Map all relevant text to the JSON Schema.
2. Output in two sections:
   - <think>: Reasoning
   - <answer>: Filled JSON

### STRICT RULES:
1. Provide both <think> and <answer>. 
   - If minimal reasoning, say: \"Direct mapping from text to schema.\"
2. Map text exactly to the JSON Schema (no omission/alteration).
3. Preserve hierarchy (ROOT $\to$ SECTION $\to$ SUBSECTION $\to$ DETAIL\_N)
4. Correctly set attributes (id, idc, idx, level\_type, component\_type, etc.).
5. JSON Format:
   - Escape quotes as \"
   - Replace newlines with \\n
   - No trailing commas
   - Only double quotes
6. Explain key decisions in <think>.

### IMPORTANT:
If <think> or <answer> is missing, response is incomplete.\"),axis=1)
\end{lstlisting}

\begin{figure}[H]
  \centering
  \includegraphics[width=0.99\textwidth]{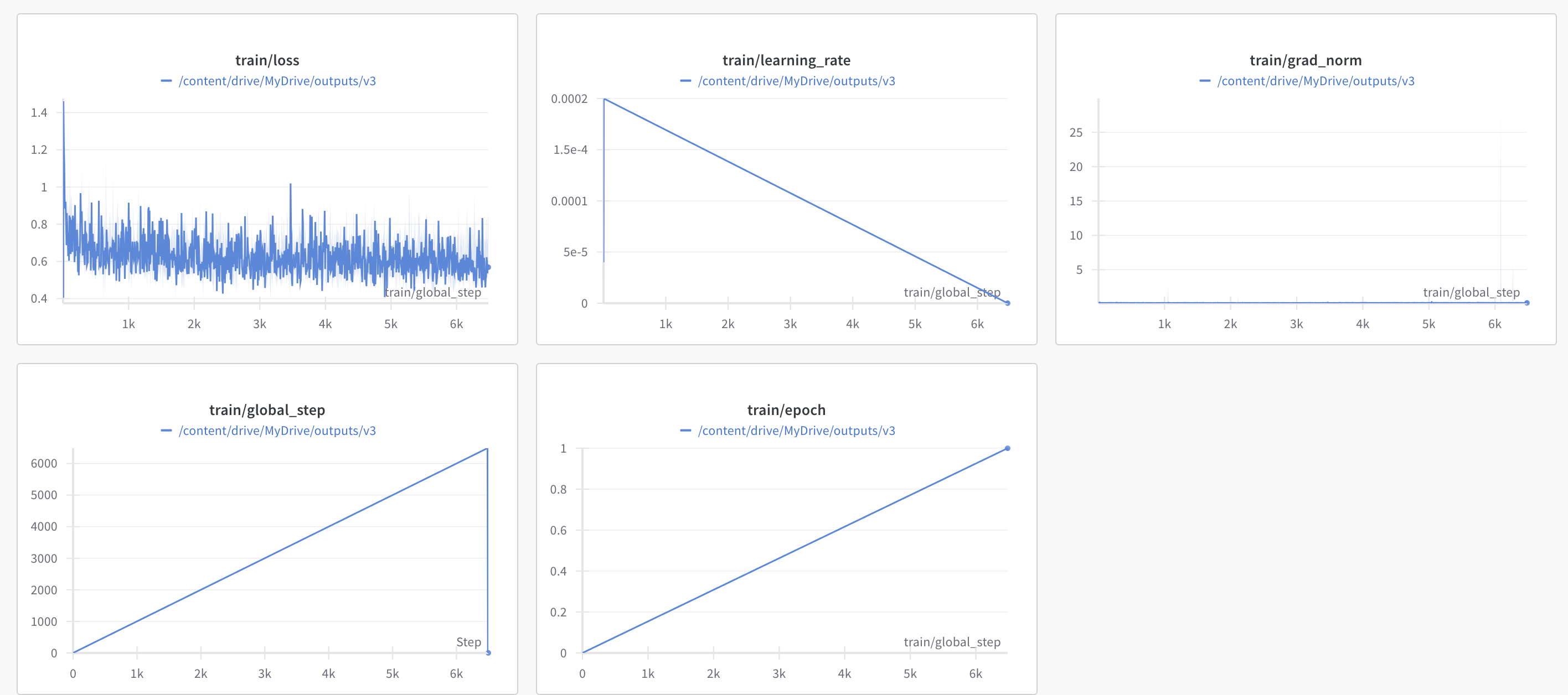}
  \caption{SFT Training Metrics}
\end{figure}

\FloatBarrier

\section{Evaluation}
We evaluated five models: ThinkJSON, Original DeepSeek R1 (671B), Distilled DeepSeek R1 (Qwen-1.5B / Qwen-7B) and Gemini 2.0 Flash (70B) which specializes on structured output generation \cite{GM2025},
 on a structured data extraction benchmark involving ~6.5K rows. Each row was processed to produce or omit a valid JSON object, and we measured metrics including:
\begin{itemize}
    \item \textbf{Rows With No Output}: Number of rows for which the model produced no structured output.
\item \textbf{Rows With Valid JSON:} Number of rows resulting in syntactically valid JSON objects.
\item \textbf{Mean Match Percentage:} Average proportion of fields correctly mapped.
\item \textbf{Mean Noise Percentage:} Average proportion of extraneous or malformed tokens within the extracted JSON.
\end{itemize}
\begin{figure}[H]
  \centering
  \includegraphics[width=0.99\textwidth]{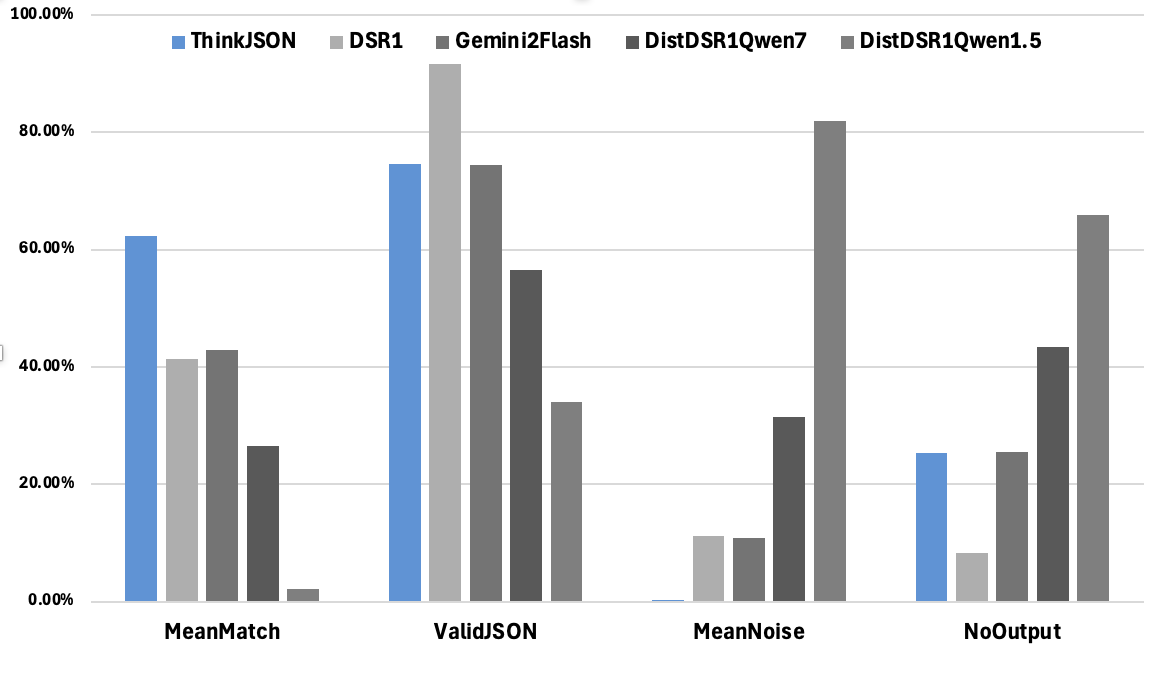}
  \caption{Performance Comparison}
\end{figure}

As illustrated, ThinkJSON yields strong results, with a 62.41\% mean match (highest of all five models) and the lowest 0.27\% mean noise, indicating minimal extraneous output. The Original DeepSeek R1 also achieves relatively high valid-JSON coverage but shows a lower mean match (41.43\%) and higher noise (11.14\%). The two distilled variants of DeepSeek R1—Qwen 1.5B and Qwen 7B—exhibit weaker performance overall, with high rates of no extracted JSON or large amounts of noise. Meanwhile, Gemini 2.0 Flash achieves a midrange mean match of 42.88\% but suffers from significant noise at 10.86\%. These findings underscore the effectiveness of our structured reasoning approach in producing concise, schema-valid outputs.

\section{Discussion and Future Direction}
Our experimental findings confirm that the reasoning-driven, schema-constrained generation pipeline is both broadly applicable—capable of handling diverse reasoning tasks beyond purely mathematical or scientific domains—and budget-conscious, as it requires comparatively moderate GPU resources and a modest dataset of reasoning examples. This balanced approach addresses a critical need in bio-manufacturing compliance , where AI systems must deliver not only correct structure but also reliable, domain-specific reasoning to meet regulatory standards \cite{NE2022}, \cite{VA2025}.

    The hallmark of our framework is integrating compliance considerations at the core of the generation process. Rather than relying on prompt-based or post-hoc solutions, our pipeline combines schema adherence objectives with iterative reasoning loops, thus reducing the need for manual oversight. This focus on strict output validation resonates with bio-manufacturing’s regulatory requirements---where precise field mappings and hierarchical consistency are crucial for electronic batch records and industry audits.

While we have employed a 1.5B-parameter foundation model, our method is readily scalable to bigger backbones (e.g., 7B parameters). Larger models could potentially yield richer context interpretation and more robust handling of rare or domain-specific phenomena. In future work, we plan to explore how increased capacity further expands the set of reasoning scenarios the model can tackle while maintaining resource efficiency—a pivotal benefit in industrial adoption.

Overall, this reinforcement + fine-tuning pipeline for structured text generation offers a flexible, compliance-aware approach that applies universal reasoning principles—spanning regulated bio-manufacturing tasks and broader domains—without incurring prohibitive computational overhead. This synergy of versatility and cost-effectiveness positions our method as a significant step forward in delivering reliable, schema-adherent AI-driven solutions.

\bibliographystyle{alpha}
\bibliography{sample}

\begin{thebibliography}{BTW23}

\bibitem[aa24]{SH2024}
Z.~Shao at~all.
\newblock Pushing the {L}imits of {M}athematical {R}easoning in {O}pen
  {L}anguage {M}odels.
\newblock {\em https://arxiv.org/pdf/arXiv:2402.03300}, 2024.

\bibitem[BTW23]{Willard2023}
Remi~Louf Brandon T.~Willard.
\newblock {E}fficient {G}uided {G}eneration for {L}arge {L}anguage {M}odels.
\newblock {\em https://arxiv.org/pdf/2307.09702}, 2023.

\bibitem[DA25]{DS2025}
DeepSeek-AI.
\newblock {D}eep{}eek-{R}1: {I}ncentivizing {R}easoning {C}apability in {LLM}s
  via {R}einforcement {L}earning.
\newblock {\em https://arxiv.org/pdf/2501.12948}, 2025.

\bibitem[ea22a]{NE2022}
Nico~Erdmann et~all.
\newblock {AI} {M}aturity {M}odel for {G}x{P} {A}pplication: {A} {F}oundation
  for {AI} {V}alidation.
\newblock {\em
  https://ispe.org/pharmaceutical-engineering/march-april-2022/ai-maturity-model-gxp-application-foundation-ai},
  2022.

\bibitem[ea22b]{Wa2022}
Yizhong~Wang et~all.
\newblock {S}elf-{I}nstruct: {A}ligning {L}anguage {M}odels with
  {S}elf-{G}enerated {I}nstructions.
\newblock {\em https://arxiv.org/abs/2212.10560}, 2022.

\bibitem[ea23a]{AM2023}
Aman~Madaan et~all.
\newblock {S}elf-{R}efine: {I}terative {R}efinement with {S}elf-{F}eedback.
\newblock {\em https://arxiv.org/abs/2303.17651}, 2023.

\bibitem[ea23b]{CD2023}
Christoph~Dann et~all.
\newblock {R}einforcement {L}earning {C}an {B}e {M}ore {E}fficient with
  {M}ultiple {R}ewards.
\newblock {\em https://proceedings.mlr.press/v202/dann23a.html}, 2023.

\bibitem[ea23c]{W2023}
P.~Wang et~all.
\newblock Math-{S}hepherd: {A} {L}abelfree {S}tep-by-{S}tep {V}erifier for
  {LLM}s in {M}athematical {R}easoning.
\newblock {\em https://arxiv.org/pdf/2312.08935}, 2023.

\bibitem[ea24a]{Shorten2024}
Connor~Shorten et~all.
\newblock {S}tructured{RAG}: {JSON} {R}esponse {F}ormatting with {L}arge
  {L}anguage {M}odels.
\newblock {\em https://arxiv.org/abs/2408.11061}, 2024.

\bibitem[ea24b]{Li2024}
Diya~Li et~all.
\newblock {L}arge {L}anguage {M}odel-{D}riven {S}tructured {O}utput: {A}
  {C}omprehensive {B}enchmark and {S}patial {D}ata {G}eneration {F}ramework.
\newblock {\em https://www.mdpi.com/2220-9964/13/11/405}, 2024.

\bibitem[ea24c]{Liu2024}
Michael~Liu et~all.
\newblock “{W}e {N}eed {S}tructured {O}utput”: {T}owards {U}ser-centered
  {C}onstraints on {L}arge {L}anguage {M}odel {O}utput.
\newblock {\em
  https://lxieyang.github.io/assets/files/pubs/llm-constraints-2024/llm-constraints-2024.pdf},
  2024.

\bibitem[ea24d]{Dong2024}
Yixin~Dong et~all.
\newblock Xgrammar: Flexible and efficient structured generation engine for
  large language models.
\newblock {\em https://arxiv.org/pdf/2411.15100}, 2024.

\bibitem[ea25a]{VA2025}
Vaishali~Aher et~al.
\newblock {A}n {O}verview on {P}harmaceutical {R}egulatory {A}ffairs {U}sing
  {A}rtificial {I}ntelligence.
\newblock {\em
  \url{https://www.ijpsjournal.com/article/An+Overview+on+Pharmaceutical+Regulatory+Affairs+Using+Artificial+Intelligence+}},
  2025.

\bibitem[ea25b]{Geng2025}
Saibo~Geng et~all.
\newblock {G}enerating {S}tructured {O}utputs from {L}anguage {M}odels:
  {B}enchmark and {S}tudies.
\newblock {\em https://arxiv.org/html/2501.10868}, 2025.

\bibitem[ea25c]{Wang2025}
Zhaoyang~Wang et~all.
\newblock {V}erifiable {F}ormat {C}ontrol for {L}arge {L}anguage {M}odel
  {G}enerations.
\newblock {\em https://arxiv.org/html/2502.04498}, 2025.

\bibitem[Lab25]{Labant2025}
MaryAnn Labant.
\newblock Smart {B}iomanufacturing: {F}rom piecemeal to all of a piece.
\newblock {\em https://www.genengnews.com/topics/bioprocessing}, 2025.

\bibitem[Sou24]{Souza2024}
Tharsis Souza.
\newblock {T}aming {LLM}s.
\newblock 2024.

\bibitem[Tea24]{QW2024}
Qwen Team.
\newblock {Q}wen. {Q}wen2.5: {A} {P}arty of {F}oundation {M}odels.
\newblock {\em https://qwenlm.github.io/blog/qwen2.5}, 2024.

\bibitem[tea25]{GM2025}
Google~AI team.
\newblock {G}enerate {S}tructured {O}utput with the {G}emini {API}.
\newblock {\em
  https://ai.google.dev/gemini-api/docs/structured-output?lang=python}, 2025.

\end{thebibliography}

\end{document}